\documentclass[letterpaper, 10 pt, journal, twoside]{IEEEtran}

\usepackage{textcomp}
\usepackage{stfloats}
\usepackage{url}
\usepackage{verbatim}
\usepackage{makecell}
\usepackage{siunitx}
\usepackage{hyperref}
\usepackage{svg}
\usepackage{graphics} 
\usepackage{graphicx}
\usepackage{amsmath} 
\usepackage{amssymb}  
\usepackage[font=footnotesize]{caption}
\usepackage[
  linesnumbered,
  noline
]{algorithm2e}
\usepackage{cite}
\usepackage[normalem]{ulem}
\usepackage[svgnames]{xcolor}
\SetAlgoCaptionLayout{centerline}
\DontPrintSemicolon

\title{Compositional Motion Generation from Demonstration\\ with Object-Centric Neural Fields
}
\author{Ahmet Tekden$^{1}$ \and Yasemin Bekiroglu$^{1,2}$
    \thanks{Manuscript received: December 18, 2025; Revised May 25, 2026; Accepted June 24, 2026.}
    \thanks{This paper was recommended for publication by Editor  Wei Pan upon evaluation of the Associate Editor and Reviewers' comments.}
    \thanks{This work was partially supported by the Wallenberg AI, Autonomous Systems and Software Program (WASP) funded by the Knut and Alice Wallenberg Foundation, the Chalmers AI Research Center (CHAIR), and the Chalmers Gender Initiative for Excellence (Genie). $^{1}$Department of Electrical Engineering, Chalmers University of Technology, SE-412 96 Gothenburg, Sweden. $^{2}$Department of Computer Science, University College London, WC1E 6BT London, U.K. Email: {\tt\small tekden@chalmers.se}}
    \thanks{Digital Object Identifier (DOI): see top of this page.}
}

\begin{document}

\maketitle

\vspace{-2em}
\begin{abstract}
Compositionality, by organizing complex behavior as combinations of simpler elements, enables robot learning that is scalable and data efficient. Leveraging this principle, we propose a generative learning-from-demonstration framework that enables compositional modeling of robotic behavior by connecting perception and motion through shared object-level representations. We render scenes from object-centric neural representations that integrate canonical neural fields with latent-conditioned deformations, capturing positional and geometric variations in a smooth, consistent, and interpretable way. For motion generation, a temporal mixture-of-experts (MoE) employs a gating mechanism to combine object-conditioned movement primitives over time, producing complete trajectories. This spatial–temporal compositionality maintains the data efficiency of movement primitives while grounding motion in visual structure, enabling systematic generalization across diverse scene configurations. In simulation, long-horizon manipulation tasks are successfully completed using the proposed model, which requires significantly less training data than other image-based baselines. Real-world experiments further demonstrate the method’s robustness to noise, its ability to generalize at the category level through language-based segmentation models, and its capacity to operate directly on 3D scene representations.

\end{abstract}

\vspace{-0.8em}
\section{Introduction}
\begin{figure}[!t]
  \centering
  \includegraphics[width=0.72\linewidth]{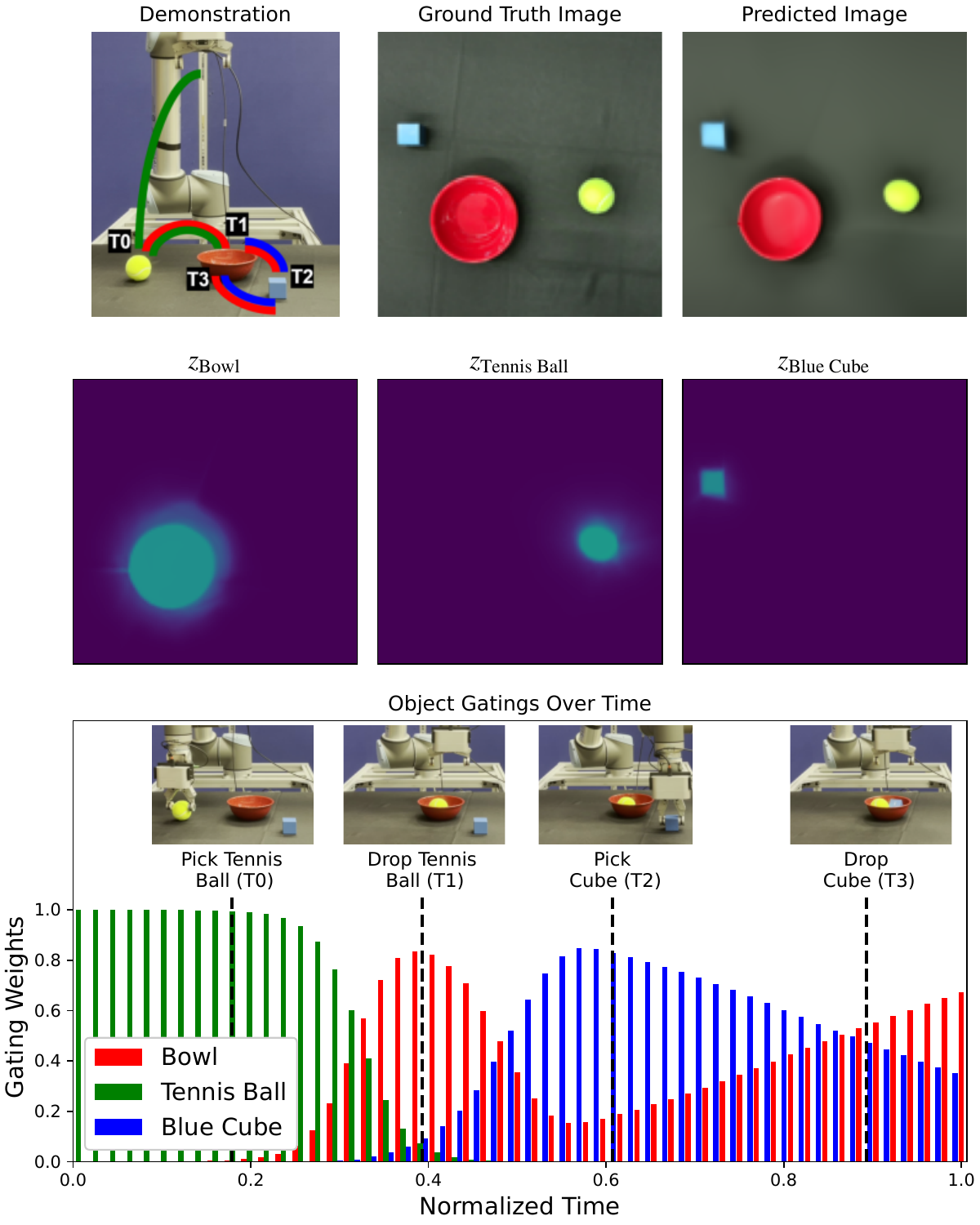}
  \caption{System overview. Top: Task sequence with object interactions, followed by the ground-truth and reconstructed images. Middle: Object-specific masks inferred from latent codes, illustrating object-centric scene decomposition. Bottom: Temporal gating weights showing each object’s influence over the trajectory; snapshots above the plot correspond to key events (T0–T3), marked by vertical dotted lines.}
  \label{fig:first-page}
\end{figure}

\IEEEPARstart{L}{}earning from Demonstration (LfD)~\cite{schaal1996learning} is a widely used approach for teaching robots task-specific skills from human demonstrations. A central challenge is representing diverse, long-horizon behaviors from a limited number of demonstrations. Treating motion as a single, monolithic sequence overlooks its compositional structure, which complicates the learning process. Everyday tasks unfold through sequential, object-level interactions—reaching, grasping, moving, and placing—that together accomplish the overall goal. This motion-level compositionality~\cite{burridge1999sequential} can be complemented by scene-level compositionality, grounding motion in object-centric scene representations and providing a natural foundation for compositional motion modeling.  Figure~\ref{fig:first-page} illustrates an example of such object-centric structure, showing how the scene is decomposed into object-specific representations and how those representations relate to different phases of a manipulation task. By supporting data efficiency, modularity, and generalization, compositional formulations present a compelling alternative to monolithic approaches~\cite{du2024position}.

Movement primitives (MPs)~\cite{dmp, promp, cnmp, vmp, li2023prodmp, tekden2023neural, yildirim2024conditional} are widely used to encode and generalize demonstrated skills. MPs provide a compact and smooth representation of trajectories from only a few demonstrations by explicitly incorporating temporal variables into motion modeling. Typically, they are employed compositionally to represent atomic skills, which can then be combined to model more complex behaviors. However, this compositionality is often imposed manually, e.g., by concatenating primitives or specifying splits with via-points. Moreover, MPs generally rely on hand-designed, low-dimensional features (e.g., object positions, goal locations). Identifying suitable features and estimating them consistently across tasks is challenging. Alternatives include learning features directly from images~\cite{cnmp}, but this typically requires large datasets and sacrifices the data-efficiency advantage of MPs.

To address these limitations, we propose a generative approach to model scenes with smooth object-level parameterizations~\footnote{Project page: \url{https://fzaero.github.io/compositional/}}. This enables task representations based on compact latent parameters that preserve the efficiency and compositionality of MPs while grounding them in perception. Building on this perspective, we propose a Mixture-of-Experts (MoE)–based motion generation framework that integrates perception and action. A spatial MoE captures smooth, object-centric generative representations that provide the building blocks for motion modeling. Object-centric structure is encouraged by soft mask–based importance sampling, ensuring that each expert learns independently without collapsing onto the same object; because masks are used only for sampling, they do not need to align precisely with object boundaries. Trajectories are represented compositionally through temporal gating, where multiple experts can be active within a segment, reflecting the relational nature of manipulation (e.g., a transfer motion depending on both grasp and place poses). Together, this spatial–temporal MoE integrates perception and action, preserves the efficiency of primitive-based methods, and enables systematic generalization to novel scenes.

The contributions of this work are as follows. A scene modeling approach is proposed that learns smooth and interpretable object-centric representations, paired with a motion modeling approach that enables sequential compositionality through temporal gating across multiple experts. Building on these approaches, a data-efficient LfD framework is introduced that integrates compositional scene and motion modeling to learn movement primitives directly from visual input, supporting generalization from few demonstrations. The framework is validated in both simulation and real-world experiments, demonstrating improved efficiency and generalization compared with existing baselines.

\vspace{-0.8em}
\section{Related Work}

Movement primitives (MPs) are a dominant representation in LfD~\cite{schaal1996learning}, providing smooth, low-dimensional encodings of trajectories. Extensions include probabilistic~\cite{promp, li2023prodmp}, task-parameterized~\cite{calinon2016tutorial, cnmp, yildirim2024conditional}, and via-point–based formulations~\cite{vmp}. Despite their effectiveness, compositionality is usually imposed manually, and MPs typically rely on hand-crafted features. Neural field–based formulation has also been explored for scene- and motion-level modeling~\cite{tekden2023neural}, but this approach operates at the scene level and scales poorly with the number of objects or scene parameters. Our framework addresses these limitations by modeling scenes and motions in a generative and compositional manner, thereby preserving the data efficiency of MPs. We further demonstrate category-level trajectory generation using object-level segmentation.

Recent imitation-learning and generative policy approaches, including diffusion-based and transformer-based methods~\cite{chi2025diffusion,zhao2023learning}, have demonstrated strong performance in robotic manipulation. More recently, Vision-Language-Action (VLA) models and generalist robot policies have explored large-scale cross-task visuomotor learning using pretrained vision-language representations and large robot datasets~\cite{zitkovich2023rt,ma2025vision}. Several recent works further investigate object-centric and few-shot adaptation strategies for pretrained VLAs, such as ControlVLA~\cite{li2025controlvla}, which combines online segmentation with pretrained visuomotor policies for object-conditioned manipulation. In contrast, our work focuses on few-demonstration, task-specific trajectory generation from a single static scene observation, where object-centric latent representations are inferred from the scene and used to condition full trajectory generation. Our experiments show that this formulation achieves high task performance given limited demonstrations.

Unsupervised scene decomposition methods, including Slot Attention~\cite{locatello2020object} and neural field–based approaches such as GIRAFFE~\cite{niemeyer2021giraffe}, learn object-centric scene representations from images. However, these methods typically rely on large-scale image datasets and are not designed for data-efficient robotic learning. In contrast, our framework learns object-centric representations from substantially fewer images and couples them with compositional motion modeling.

Neural fields~\cite{nerf_survey} provide powerful continuous representations and have been extended beyond reconstruction~\cite{nerf,deepsdf,occ-net} to robotics tasks. For motion generation, prior work has applied them to correspondence~\cite{NDF, tekden2023grasp}, trajectory-level modeling~\cite{tekden2023neural} and reactive motion generation~\cite{tekden2026reactive}. However, existing approaches either model motion only at the waypoint or pose level, focus on motion generation without supporting visual conditioning or sacrifice object-level compositionality and data efficiency. In contrast, we introduce a neural field–based framework for compositional scene and motion modeling at the object level, enabling efficient full-trajectory generation from visual input.

\vspace{-0.8em}
\section{Problem Formulation}

We focus on learning to generate motion from demonstrations, leveraging compositional representations of scenes and motion to enable generalization from a few examples. The training data consist of a set of images with coarse object masks and corresponding motion trajectories:
\begin{equation}
\mathcal{D} = \left\{ \big(I^{(k)}, \{M_i^{(k)}\}_{i=1}^{N}, \tau^{(k)} \big) \right\}_{k=1}^{K},
\end{equation}
where $I^{(k)}$ is the $k$-th image, $\{M_i^{(k)}\}_{i=1}^{N}$ are the object masks for $N$ objects, 
and $\tau^{(k)} = \{q^{(k)}(t)\}_{t=1}^T$ is motion trajectory of length $T$, where $q^{(k)}(t)$ denotes the robot state at time $t$, such as the end-effector position. The dataset includes demonstrations that cover the boundary values of the task parameters (e.g., the minimum and maximum admissible object configurations). This ensures that the model observes the full range of object variability present in the task. Each object mask $M_i^{(k)}$ marks the image region of a single object. The masks should sufficiently cover the target object, but do not need to precisely align with object boundaries~\footnote{In practice, the masks are annotated using one or two rectangles.}. We assume that objects occupy distinct regions and that the background is consistent across demonstrations during training, including a fixed camera viewpoint and consistent scene layout. This makes annotation easier and ensures that image variations reflect object-level changes.

We assume trajectories are approximately time-aligned across demonstrations, so that each index $t$ corresponds to the same task phase (e.g., approach, grasp, transport, release). Such trajectories can be obtained through motion-planned demonstrations or by aligning kinesthetic trajectories based on contact events or changes in motion profiles~\cite{calinon2016tutorial}. This establishes a shared temporal structure across demonstrations and enables sequential compositionality, where different object latent representations influence distinct trajectory segments.

The learning objective is (i) to learn a scene representation model along with latent vectors $\{z_i\}_{i=1}^N$ for each object such that each scene can be expressed as a composition of a fixed background and objects whose variations (e.g., in position or geometry) are captured smoothly in the latent space; and (ii) to train a motion generation model that learns to map these latent vectors to the demonstrated trajectories. In the inference phase, latent vectors for a new scene are extracted using the learned scene representation model and fed into the motion model to generate the corresponding trajectory.

\vspace{-0.8em}
\section{Method}

\begin{figure}[t]
  \centering
  \includegraphics[width=0.75\linewidth]{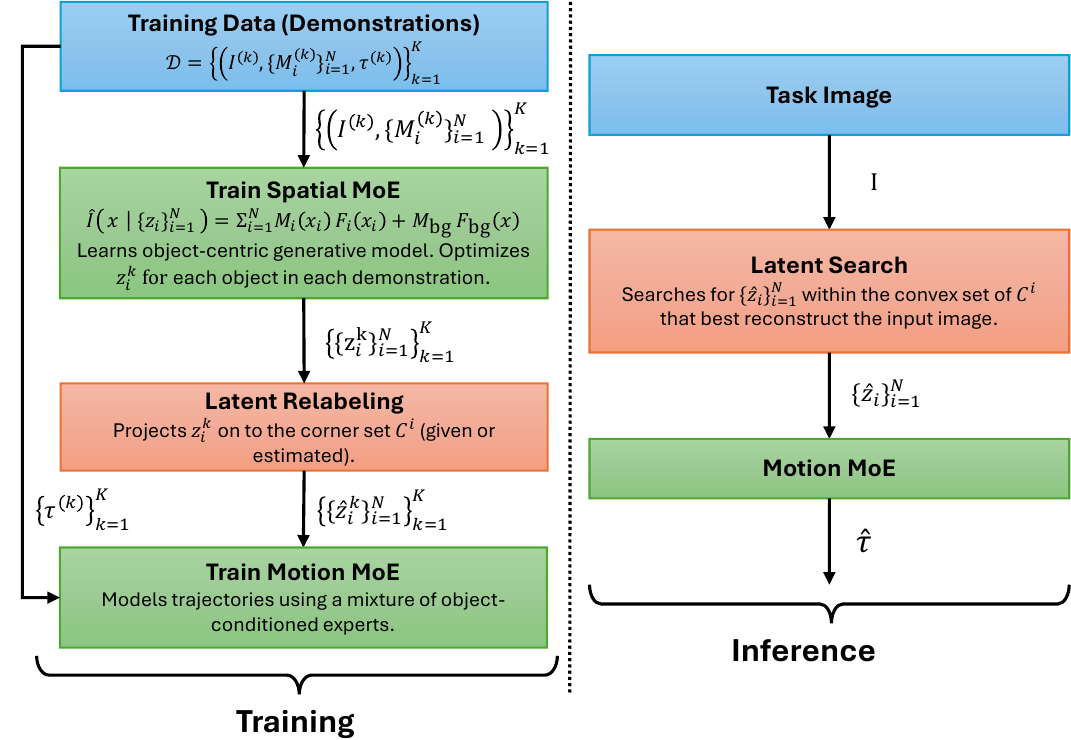}
  \caption{Overview of training and inference pipelines. The Spatial MoE learns object-centric latent vectors, which are canonicalized via latent relabeling and used to train the Motion MoE. During inference, object latents recovered via latent search are used by the Motion MoE to generate trajectories.
  }
  \label{fig:arch}
\end{figure}

The proposed framework is shown in Fig.~\ref{fig:arch}. We first learn an object-centric generative model from demonstration images (Sec.~\ref{sec:compositional-image-modeling}, Sec.~\ref{sec:object-centric-image-modeling}), yielding latent representations that capture each object in the scene. We then construct a minimal canonical latent set that can equivalently reconstruct the training images (Sec.~\ref{sec:latent-relabeling}). Finally, we model motion compositionally by conditioning a temporal MoE on these canonical latent representations (Sec.~\ref{sec:motion-modeling}), enabling trajectory generation grounded in object-centric structure.

\vspace{-0.8em}
\subsection{Compositional Image Modeling}\label{sec:compositional-image-modeling}

\begin{figure}[t]
    \centering
    \includegraphics[width=1\linewidth]{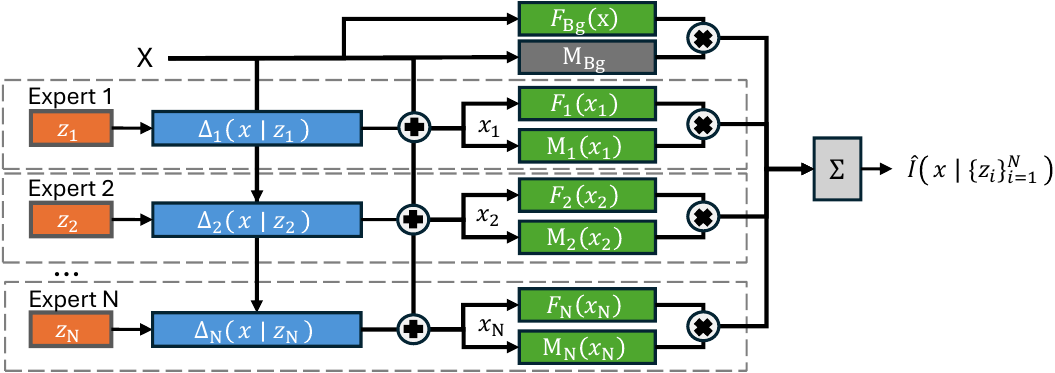}
    \caption{Given N expert-specific objects, orange denotes learnable embeddings, blue corresponds to FiLM-conditioned MLPs, and green represents standard MLPs. Light gray indicates mathematical operators, and dark gray denotes the background mask computed from a constant, non-trainable logit.}
    \label{fig:Image-Moe}
\end{figure}

We model each image using an object-centric spatial MoE:
\begin{equation}
\hat I(x| \{z_i\}_{i=1}^N) = \sum_{i=1}^{N} M_i(x_i)\,F_i(x_i) + M_{\text{bg}}\,F_{\text{bg}}(x),
\label{eq:image_model}
\end{equation}
with deformed coordinates $x_i = x - \Delta_i(x \mid z_i),$ where $F_i(x)$ predicts the appearance of object $i$, $\Delta_i(x\mid z_i)$ is the deformation conditioned on latent vector $z_i$, and $M_{\text{bg}}, F_{\text{bg}}$ denote the background mask and appearance model, respectively. All fields $F_i, F_{\text{bg}}$, and deformation $\Delta_i$ are implemented as coordinate-based multilayer perceptrons (MLPs), i.e., neural fields. The spatial MoE is shown in Figure~\ref{fig:Image-Moe}.

Object masks are computed via a spatial softmax across objects and background:
\begin{equation}
M_i(x_i) = \frac{\exp\!\left(l_i(x_i)\right)}{\exp(l_{\text{bg}}) + \sum_{j=1}^N \exp\!\left(l_j(x_j)\right)},
\end{equation}
where the background logit $l_{\text{bg}}$ is constant, and each object logit is given by a learned function $l_i(x_i)$. The logit functions are also implemented as coordinate-based MLPs. This formulation yields object masks that partition the image into spatially coherent regions, while deformations capture scene variations as smooth changes in object position and shape. 

\vspace{-0.5em}
\subsection{Object-Centered Training of Experts} \label{sec:object-centric-image-modeling} 
\label{training-experts}
Training the compositional image model proceeds in multiple stages. We first estimate the static background, then pretrain individual object fields using coarse masks, and finally refine all components jointly under the mixture formulation. Optimization in all stages uses an $\ell_2$ reconstruction loss between predicted and ground-truth pixels. For deformation networks $\Delta_i$, we additionally employ Lipschitz regularization~\cite{lips-reg} to encourage the mapping $z_i \mapsto \Delta_i(x \mid z_i)$ to vary smoothly with latent variables. While the neural-field parameterization already induces a generally smooth dependence on $z_i$, this regularization further helps by discouraging abrupt local changes and improving stability.

\paragraph{Background Modeling}
We first train $F_{\text{bg}}$ using only non-masked pixels from all demonstrated images. Compared to naïve median-image-based supervision, this procedure provides a more consistent background estimation.

\paragraph{Object-Centric Training}

Each object model $F_i$ is trained using its corresponding mask within an object-centric sampling framework. During training, pixels that do not belong to other objects are sampled from the demonstration image, with additional importance sampling applied to $M_i$ to emphasize the appearance of the target object. Pixels that fall within the masks of other objects are instead sampled from the background model, preventing $F_i$ from capturing regions associated with different objects. This sampling strategy ensures that variations across scenes are attributed solely to the correct object and enables the learned masks to refine coarse input masks into more precise segmentations.
  
To further improve training stability, we employ a position-aware zoom-based curriculum around each object mask. This approach is particularly beneficial for smaller objects, as it increases their effective pixel coverage and provides more stable gradients during optimization. The zoom factor is gradually reduced over time, allowing the model to converge toward accurate object representations at their true scale. 

\paragraph{Joint Compositional Training}
Finally, we continue training using the full compositional model in \eqref{eq:image_model}, with background and object models jointly optimized. The earlier stages act as pretraining, while this stage refines consistency between objects and background. Overall, this object-aware training scheme enables the model to explain variations across scenes as object-dependent differences, allowing robust generalization with as few as 10–30 training images.

\paragraph{Extension to 3D Representations} 
While we describe $F_i$ as 2D appearance fields, the mixture formulation is not restricted to images. Each $F_i$ can also be defined as an occupancy field~\cite{occ-net} over 3D space, enabling the construction of a compositional 3D scene representation.

\begin{algorithm}[t]
\footnotesize
\vspace*{2pt}
\hrule height 0.6pt
\vspace*{2pt}
\KwIn{Reference image $I$, corner set $C^i$ for object $i$, number of samples $P$, preselection size $L'$, maximum candidate count $L$, error threshold $\lambda$}
\KwOut{Top-$L$ latent vector candidates $\{\hat{z}_{i,l}^{(k)}\}_{l=1}^L$}

Sample $P$ weight vectors $\{\alpha_p\}_{p=1}^{P} \sim \mathrm{Dirichlet}(|C^i|)$\;

\For{$p = 1$ \KwTo $P$}{
    $\hat{z}_{i,p}^{(k)} = \sum_{j} \alpha_{p,j} \, c_{i,j}, \quad c_{i,j}\in C^i$\;
    \vspace*{1.5pt}
    $E_{i,p}^{(k)} = \| I - \hat{I}(\cdot \mid \hat{z}_{i,p}^{(k)}) \|_2$ 
}

Select the top-$L'$ candidates with the lowest reconstruction errors\;

Refine these candidates via gradient descent on $E_{i,l}^{(k)}$\;

$E_{\min} = \min_l E_{i,l}^{(k)}$\;
Retain all candidates satisfying $E_{i,l}^{(k)} < \lambda E_{\min}$, up to $L$ total\;

\Return $\{\hat{z}_{i,l}^{(k)}\}_{l=1}^{L}$\;
\vspace*{2pt}
\hrule height 0.6pt 
\vspace*{3pt}
\caption{Latent Search}
\label{alg:latent_search}
\end{algorithm}

\vspace{-0.8em}
\subsection{Latent Relabeling} \label{sec:latent-relabeling}
Each object is associated with a latent vector $z_i^{(k)} \in \mathbb{R}^{d_z}$ for each demonstration $k$. Smooth object variations are captured by interpolating between a small number of endpoint latent vectors that span the boundaries of the observed variation. For each object $i$, these endpoints form the corner set $C^i$. Given this set, the search (Algorithm~\ref{alg:latent_search}) recovers object-specific latent representations for each demonstration by sampling convex combinations of the corners and refining the lowest-error candidates through gradient-based optimization. As multiple convex combinations can yield similar reconstructions, we retain the top-$L$ candidates per demonstration to preserve valid alternatives for downstream motion modeling. This effectively increases the training set size from $K$ to $K\times L$.

We estimate the corner sets by iteratively expanding $C^i$. We initialize $C^i$ with two embeddings that are maximally distant within the set $\{z_i^{(k)}\}_{k=1}^K$ and use Algorithm~\ref{alg:latent_search} to test whether the remaining instances can be reconstructed from the current corner set. For any instance that cannot be reconstructed, we expand $C^i$ by selecting the embedding most distant from the latents sampled via convex combinations of the current corner set, leveraging the observation that larger visual differences are generally reflected by greater distances in the latent space. This process iteratively widens the representable region of the latent space until all training instances can be reconstructed from the corner set, ensuring that every demonstration latent can be canonicalized as a convex combination of the corners. As a result, generalization is naturally constrained to variations represented within the demonstrated latent support, and extrapolation beyond this region is not guaranteed.

\vspace{-0.5em}
\subsection{Motion Generation} \label{sec:motion-modeling}

\begin{figure}[t]
    \centering
    \includegraphics[width=0.9\linewidth]{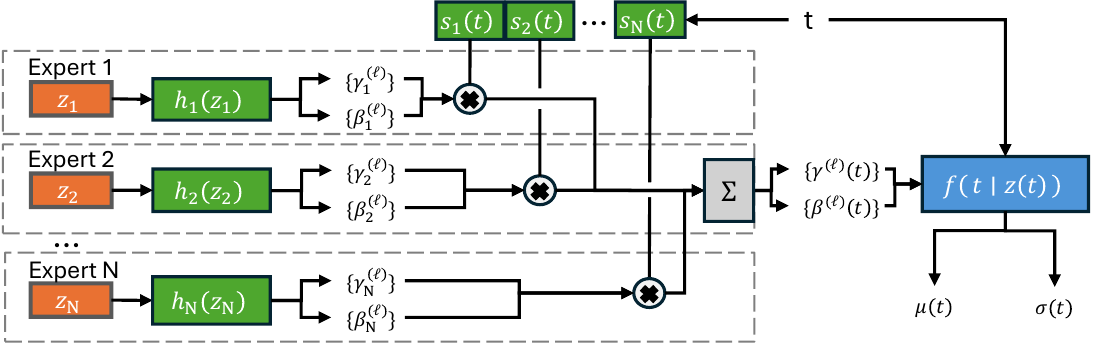}
    \caption{Given N expert-specific object embeddings, FiLM weights and object relevance scores are predicted via MLPs (green). These are pooled to generate FiLM parameters used to condition the trajectory generation.}
    \label{fig:Motion-Moe}
\end{figure}

Given the relabeled object latent vectors $\{\hat{z}_i\}_{i=1}^N$, we assume sequential compositionality: different trajectory segments depend on different subsets of $\hat{z}_i$. At an abstract level, we model this through a time-dependent conditioning signal constructed via softmax mixture weights. The exact conditioning mechanism is specified later in Sec.~\ref{sec:film}.

\begin{equation}
c(t) \;=\; \sum_{i=1}^N w_i(t)\, c_i(\hat{z}_i) , \quad w_i(t) \;=\; \frac{\exp\!\big(s_i(t)\big)}{\sum_{j=1}^N \exp\!\big(s_j(t)\big)}
\label{eq:pooled_conditioning}
\end{equation}
where $s_i(t)$ are learnable relevance scores and $c_i(\hat{z}_i)$ denotes object-specific conditioning variables. The generator then predicts a trajectory in the state space $q(t) \in \mathbb{R}^d$ as
\begin{equation}
q(t) \sim \mathcal{N}\!\left(\mu(t),\, \mathrm{diag}\!\big(\sigma^2(t)\big)\right).
\end{equation}
where $\mu(t)$ and $\sigma(t)$ are estimated via 
\begin{equation}
[\mu(t),\, \sigma(t)] = f(t \mid c(t)),  
\label{eq:traj_dist}
\end{equation}
where $\sigma(t)$ is constrained to be positive via a softplus transform applied to the network output. The motion generation framework is shown in Figure~\ref{fig:Motion-Moe}.

The model parameters are optimized by minimizing the negative log-likelihood of the demonstrated trajectories, where each trajectory is modeled as a Gaussian:
\begin{equation}
\mathcal{L}_{\text{traj}} 
= - \sum_{k,t} 
\log \mathcal{N}\!\big(q^{(k)}(t)\,;\, 
\mu(t \mid c(t)),\, \Sigma(t \mid c(t)) \big),
\label{eq:traj_loss}
\end{equation}
where $\Sigma(t \mid c(t)) = \mathrm{diag}(\sigma^2(t \mid c(t)))$. To discourage the model from learning spurious dependencies on irrelevant latent variables, we apply latent dropout during early training by randomly masking a subset before pooling. This warm-up regularization encourages the generator to rely primarily on informative variables, while in later training and at test time, all variables are used. Additionally, the probabilistic trajectory formulation mitigates overfitting to spurious correlations by allowing irrelevant latent variables to be absorbed into the variance term rather than biasing the mean.

\paragraph{Modeling Orientation} 
For modeling orientation, we follow~\cite{ude_quat} and represent the quaternion trajectories $q_{1:T}$ in axis–angle form $r_{1:T}$, expressed in the tangent space of the initial quaternion $q_1$, yielding continuous rotation trajectories that can be modeled similarly to Cartesian trajectories. 

\subsection{Latent Conditioning with FiLM} \label{sec:film}

Both the image model and the trajectory generator are conditioned on latent vectors. We implement this conditioning through \emph{Feature-wise Linear Modulation (FiLM)}~\cite{perez2018film}, which reconfigures the parameters of a base network using latent-dependent scale and shift terms. For a given $z_i$, FiLM produces parameters
\begin{equation}
h_i(z_i) = \{ \gamma_i^{(\ell)}, \, \beta_i^{(\ell)} \}_{\ell=1}^L,
\end{equation}
\\
\noindent 
where $\gamma_i^{(\ell)}, \beta_i^{(\ell)} \in \mathbb{R}^H$ for layer width $H$. Given base layer parameters $(w^{(\ell)}, b^{(\ell)})$, the modulated parameters are
\begin{equation}
\tilde w^{(\ell)} = \gamma^{(\ell)} \odot w^{(\ell)}, \quad
\tilde b^{(\ell)} = \gamma^{(\ell)} \odot b^{(\ell)} + \beta^{(\ell)},
\end{equation}
with $\odot$ denoting elementwise multiplication. FiLM thus provides a lightweight mechanism to adapt a shared architecture to different objects or tasks. This role is analogous to hypernetworks~\cite{hypernet}, which generate full network weights from latent codes, but FiLM achieves similar conditioning capability with fewer parameters and improved stability.

In the compositional image model, FiLM parameters are derived from each $z_i$ and applied directly when predicting object deformations. In the trajectory generator, the conditioning variables in \eqref{eq:pooled_conditioning} correspond to the FiLM modulation parameters, i.e., $c_i(\hat z_i) \equiv \{\gamma_i^{(\ell)}, \beta_i^{(\ell)}\}_{\ell=1}^L.$ Accordingly, the aggregated conditioning signal is implemented through
\begin{equation}
\gamma^{(\ell)}(t) = \sum_{i=1}^N w_i(t)\, \gamma_i^{(\ell)}, \quad
\beta^{(\ell)}(t) = \sum_{i=1}^N w_i(t)\, \beta_i^{(\ell)} .
\end{equation}
This mechanism realizes sequential compositionality: different trajectory segments are influenced by different subsets of $z_i$, depending on which $w_i(t)$ are active.

\vspace{-0.5em}
\subsection{Inference}

At test time, we apply the same latent search procedure (Algorithm~\ref{alg:latent_search}) using the corner sets from training. 
Objects are processed sequentially, with each contribution added incrementally to condition subsequent reconstructions. In rare cases where visually similar objects (e.g., of the same color) lead to ambiguity, larger objects are prioritized to provide more reliable reconstruction cues. Since reconstruction is based on visual similarity to the learned object representations, the method tolerates moderate lighting variation and visually distinct distractor objects. During latent inference, representations that better reconstruct the learned task-relevant object structure are favored, allowing distractor objects that are not well represented by the learned object model to be ignored. However, distractors that are visually similar to task-relevant objects may still lead to ambiguity or degraded performance. This procedure recovers object-specific latent representations for novel task instances while remaining within a compact and well-structured region of the latent space. The resulting latent representations then condition the trajectory generator. In the current implementation, running on an Intel(R) Xeon(R) W-2245 CPU @ 3.90GHz and an NVIDIA Quadro RTX 6000 GPU, inference takes approximately 1.4 and 0.8 seconds per object for RGB and 3D settings, respectively~\footnote{3D inference employs sparsely sampled coordinates near surfaces instead of the full coordinate space.}.

\section{Experiments}

\begin{figure}[!t]
    \centering
    \includegraphics[width=0.95\linewidth]{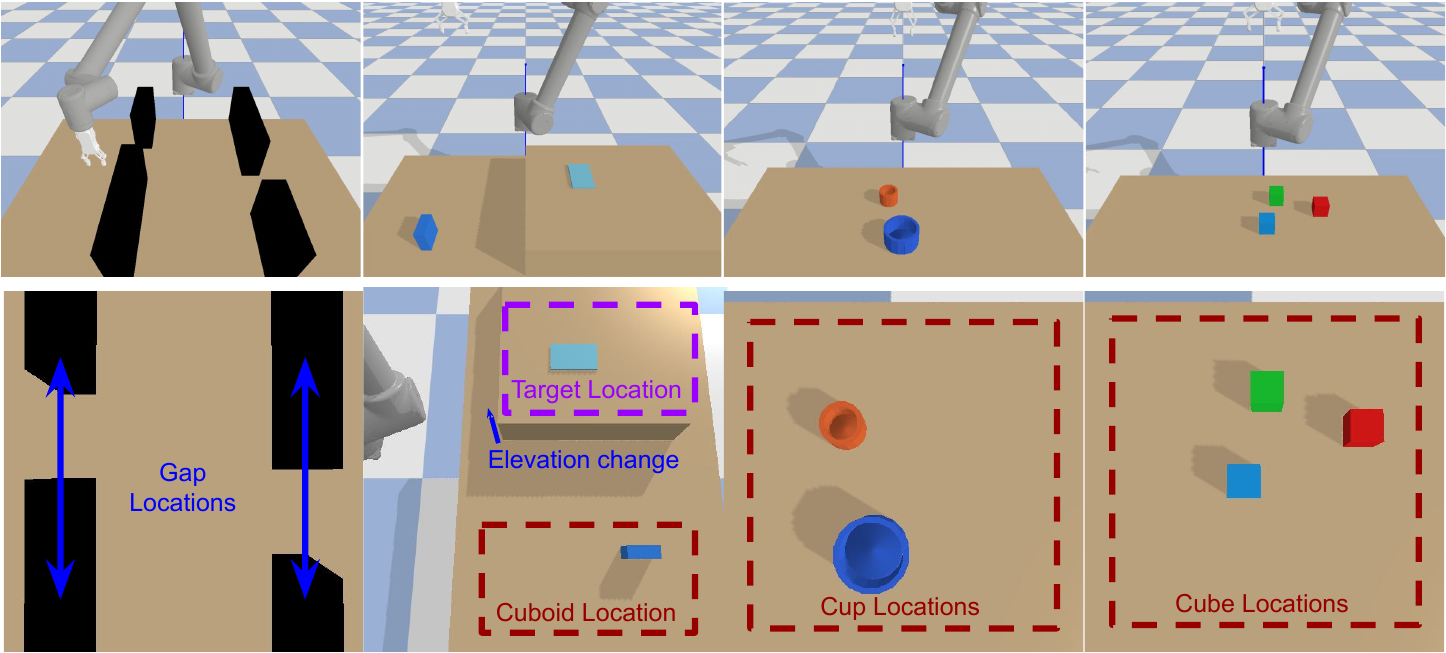}
    \caption{Simulation tasks with corresponding scene variations. The bottom row shows the camera images used by the methods. In these images, arrows highlight geometric changes and rectangles mark possible object locations.}
    \label{fig:sim_tasks}
\end{figure}

We evaluate the framework through simulation benchmarks and real-world robotic experiments. The simulation studies show how compositional motion modeling through object-centric latent representations and time-varying latent aggregation improves generalization and data efficiency. Real-world experiments demonstrate the method’s ability to handle perception noise, integrate 3D scene representations, and enable category-level generalization through vision-language–based segmentation.

\subsection{Simulation Experiments}

We first assess the proposed method on four manipulation tasks of increasing complexity. In Wall Avoidance (easy), the robot reaches a goal by navigating through openings in two walls. Incline Pick-and-Place (medium) requires transporting a cuboid onto a platform at varying elevations, with both the cuboid and target locations changing across trials. Cup Stacking (medium) involves inserting a small cup into a larger one, while Cube Stacking (hard) requires stacking three cubes from diverse initial positions. These tasks introduce scene variations including geometric changes and varying object placements, as illustrated in Figure~\ref{fig:sim_tasks}.

We compare the proposed method against several baselines, including conditional neural movement primitives (CNMP)~\cite{cnmp} with CNN and Diffusion Policy (DP)~\cite{chi2025diffusion}. CNMP supports image-conditioned trajectory generation, making it one of the closest comparable approaches for our setting. DP is a recent imitation-learning framework that has demonstrated strong performance in robotic manipulation~\footnote{To align the prediction setting across methods, normalized time is provided as a state input to DP.}. To evaluate the proposed perception pipeline, we include CNN-FiLM and DINO-FiLM ablations that replace the object-centric latent representation with global image features. CNN-FiLM uses end-to-end learned CNN features, while DINO-FiLM uses pretrained DINO~\cite{dinov3} features. We further include an ablation without temporal gating (“Ours w/o gating”) to isolate the contribution of time-varying latent aggregation. Finally, we include a modified Neural Field Movement Primitive baseline (NFMP)~\cite{tekden2023neural}, where we replace the hypernetwork with FiLM layers and omit positional encodings for a fairer comparison. Since NFMP relies on grid-based training over all combinations of task parameters, scaling as $3^n$ with $n$ task parameters, it is evaluated only on the Wall Avoidance task, where the required number of demonstrations remains tractable.

\begin{table}[!t]
\vspace{3mm}
    \centering
    \caption{Simulation results (MED/Accuracy).}
    \label{tab:sim_results}\renewcommand{\arraystretch}{1.2}
    \setcellgapes{2pt} \makegapedcells
    \setlength{\arrayrulewidth}{0.3pt}
    \resizebox{\columnwidth}{!}{
    \begin{tabular}{|l|cccc|}
\hline
Method (data regime) & 
\makecell{Wall\\Avoidance} & 
\makecell{Incline\\Pick-and-Place} & 
\makecell{Cup\\Stacking} & 
\makecell{Cube\\Stacking} \\
\hline
 CNN-CNMP (Low)             & 2.72±0.08/0.67±0.04                   & 4.35±0.27/0.11±0.03                   & 10.85±1.39/0.02±0.01                  & 14.22±0.59/0.00±0.00                  \\
 CNN-CNMP (High)            & 1.70±0.26/0.95±0.02                   & 2.39±0.43/0.51±0.15                   & 3.08±0.27/0.16±0.07                   & 6.07±0.32/0.00±0.00                   \\
  DP (Low)                   & 2.31±0.04/0.79±0.02                   & 3.60±0.08/0.25±0.01                   & 11.13±1.29/0.02±0.02                  & 11.55±0.19/0.00±0.00                  \\
 DP (High)                  & 0.55±0.03/\textbf{0.99±0.00}          & 1.49±0.14/0.86±0.04                   & 2.72±0.09/0.26±0.04                   & 4.77±0.61/0.11±0.03                   \\
 NFMP (Grid)                & 0.85±0.18/0.98±0.01                   & NA                                    & NA                                    & NA                                    \\
 \hline
 DINO-Film (Low)            & 4.46±0.19/0.65±0.02                   & 5.03±0.13/0.21±0.00                   & 9.81±0.34/0.00±0.00                   & 15.10±0.07/0.00±0.00                  \\
 DINO-Film (High)           & 2.27±0.12/0.84±0.00                   & 3.71±0.06/0.31±0.06                   & 6.69±0.06/0.01±0.00                   & 10.48±0.08/0.00±0.00                  \\
 CNN-Film (Low)             & 1.41±0.19/0.89±0.02                   & 3.01±0.55/0.40±0.13                   & 8.74±1.83/0.03±0.01                   & 9.26±0.74/0.00±0.00                   \\
 CNN-Film (High)            & \textbf{0.46±0.09}/\textbf{0.99±0.00} & 1.18±0.09/0.92±0.05                   & 2.11±0.20/0.26±0.06                   & 1.32±0.03/0.66±0.10                   \\
  \hline
 Ours w/o gating (Low)      & 0.71±0.04/\textbf{0.99±0.00}          & 0.89±0.03/0.99±0.01                   & 1.10±0.03/0.86±0.05                   & 1.33±0.05/0.85±0.06                   \\
  \hline
 Ours (Low)                 & \textbf{0.51±0.03}/\textbf{0.99±0.00}          & \textbf{0.71±0.02}/\textbf{1.00±0.00} & \textbf{0.65±0.04}/\textbf{0.99±0.00} & \textbf{0.72±0.05}/\textbf{0.97±0.02} \\
\hline
\end{tabular}

}
\begin{flushleft}
\scriptsize
\hangindent=0.5em
\hangafter=0
Low/High training sizes per task: Wall Avoidance (10/30), Incline Pick-and-Place (20/50), Cup Stacking (20/50), and Cube Stacking (30/100). Bold numbers indicate the best performance in each task under the low- and all-data regimes, respectively. Lower MED and higher accuracy indicate better performance.
\end{flushleft}
\end{table}
\begin{figure}[!t]
    \centering
    \includegraphics[width=1\linewidth]{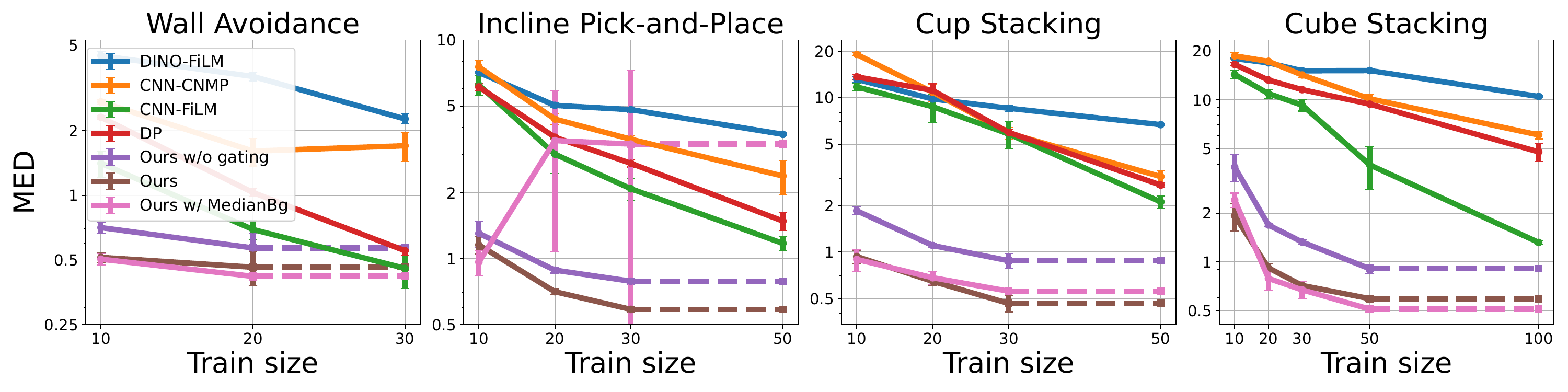}
    \caption{Training size versus MED, illustrating the data efficiency of the proposed approach. The proposed method achieves low error with far fewer demonstrations than the baselines, particularly on more challenging tasks.}
    \label{fig:med-ts}
\end{figure}

Performance is evaluated using three criteria: mean Euclidean distance (MED) between predicted and demonstrated trajectories, task success rate, and data efficiency across varying training sizes. All models are trained with three random seeds, and results are reported as the mean and standard deviation across seeds. We consider two data regimes: a low-data regime with limited demonstrations per task, and a high-data regime with a larger demonstration set. Baselines are evaluated in both regimes, whereas the proposed method is evaluated only in the low-data regime, demonstrating that it matches or outperforms baselines trained with substantially more demonstrations. 

Results are presented in Table~\ref{tab:sim_results}. Across tasks, the proposed method, both with and without object-centric temporal gating, consistently outperforms baselines in the low-data regime. The CNN-FiLM baseline generally achieves higher performance than CNN-CNMP and Diffusion Policy, suggesting that FiLM-conditioned trajectory generation itself provides a strong inductive bias in our setting. Even when trained with many more demonstrations, baseline performance is generally comparable to, or lower than, that of the proposed model. Comparing CNN-FiLM with the proposed method without temporal gating further highlights the contribution of the proposed object-centric representation. The only exception is the Wall Avoidance task, where CNN-FiLM achieves a marginally lower MED when trained with three times as many demonstrations. However, both methods successfully avoid the wall, and the relatively low task complexity reduces the data-efficiency advantage of the proposed method.  To further illustrate data efficiency, Figure~\ref{fig:med-ts} shows MED as a function of training size, where each point represents the mean and standard deviation across seeds. The proposed method achieves consistently low error using few demonstrations, while baselines require significantly more data to reach similar performance, particularly on more challenging tasks. The figure also includes an ablation where $F_{bg}$ is trained using median-image-based supervision. While the average performance remains broadly comparable, the median-image-based variant occasionally exhibits less stable behavior and larger error cases.

Finally, we evaluated robustness to incomplete masks and imperfect temporal alignment on the Cube Stacking task across three seeds. The method maintained MED comparable to the original setup when up to 3 out of 30 demonstrations contained incomplete masks, while larger numbers prevented reliable object decomposition during scene-model training. For temporal alignment, we introduced temporal warping within sub-task segments while preserving key task phases. This increased the mean MED to $2.03$ and reduced the task success rate to approximately $0.90$, suggesting that the method retains reasonable performance under moderate intra-phase timing variation while still relying on coarse phase alignment across demonstrations.

\begin{figure}
    \centering
    \includegraphics[width=0.90\linewidth]{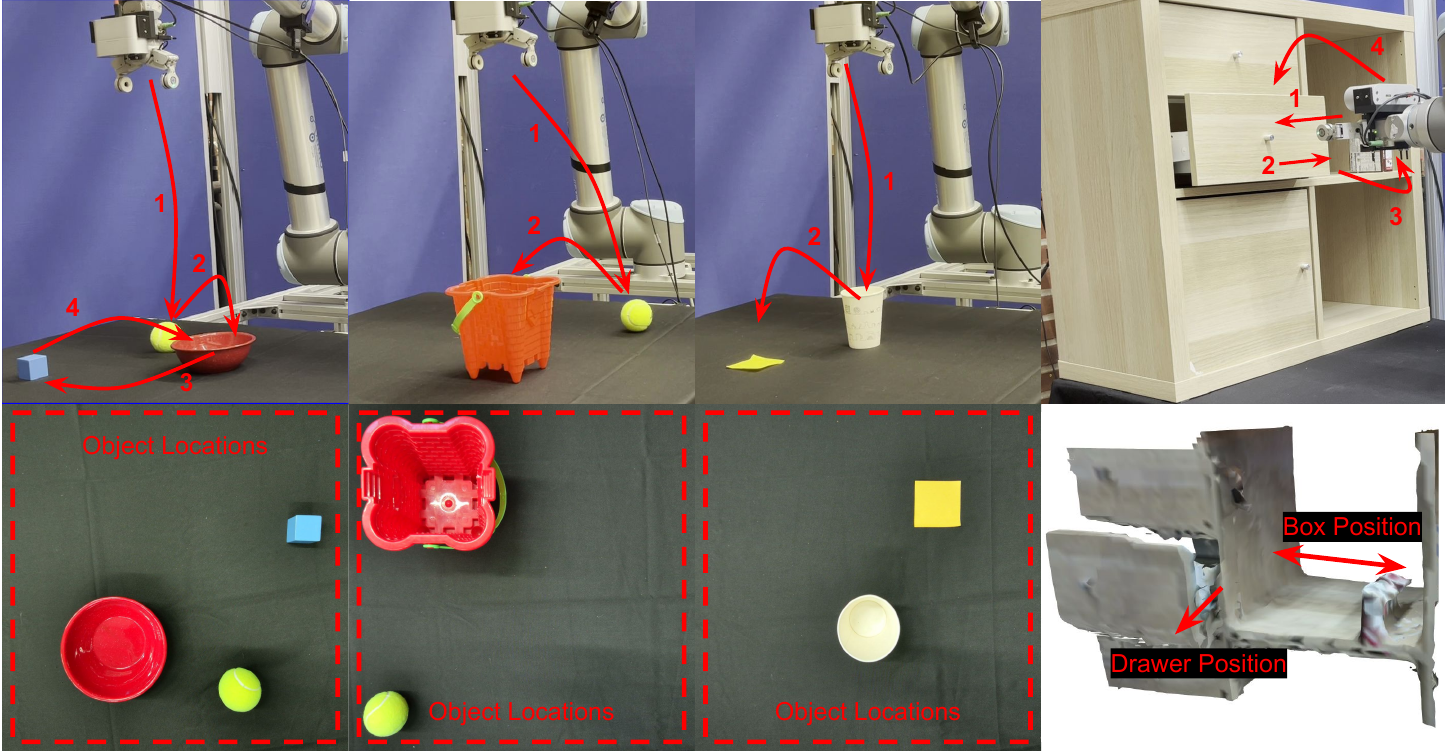}
    \caption{Real-world task and scene variations. The top row illustrates the task steps, and the bottom row provides sample inputs, with rectangles and arrows marking possible scene parametrization. As occupancy values are hard to visually discern, we show the underlying mesh for the final task.}
    \label{fig:real-world-tasks}
\end{figure}

\vspace{-1mm}
\subsection{Real-Robot Experiments}

Validation of the framework is performed on four real-world robot tasks. These experiments demonstrate the method’s robustness to perception noise, its ability to generalize across object categories using segmentation masks obtained from vision-language models, and its compatibility with 3D scene representations. These tasks are shown in Figure~\ref{fig:real-world-tasks}. The demonstrations are six-degree-of-freedom trajectories collected via kinesthetic teaching, except for the fourth task, where motion planning is used due to the complexity of the required movements. All demonstration trajectories are temporally aligned using gripper-state transitions (open/close), which serve as consistent semantic markers across demonstrations. During testing, we simplify motion execution by segmenting the predicted trajectories at gripper-state transitions (detected via rising and falling edges). Each sub-trajectory is then executed sequentially, and the corresponding gripper command is issued at the transition point. Trajectory execution is performed using MoveIt~\cite{moveit}. A small validation set is used for early stopping during the training of the motion generation model. 

\begin{figure}
    \centering
    \includegraphics[width=0.95\linewidth]{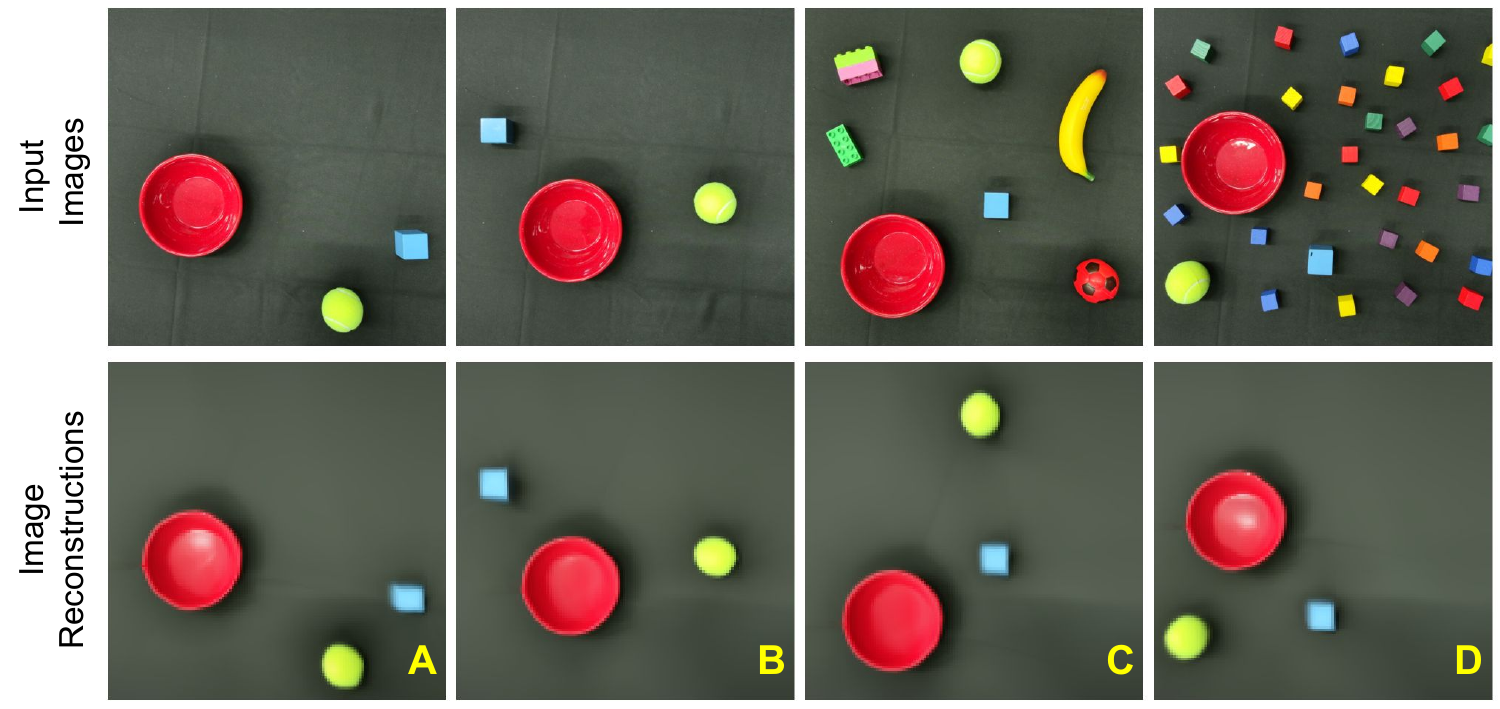}
    \caption{Input Images, and corresponding image reconstructions with (C, D) and without (A, B) distractor objects for the first task.}
    \label{fig:task1-preds}
\end{figure}

In the first task, the robot picks up tabletop objects and places them in a bowl, with varying locations across trials. It requires approximately 1–2 cm horizontal grasping accuracy to reliably grasp the tabletop objects. With 30 training demonstrations, the system successfully completes all 25 test cases~\footnote{With 20 training demonstrations, it completes 3 out of 5 additional trials. Failures occur when grasping the blue cube, which requires higher precision.}. We further evaluate robustness by adding visually distinct distractor objects in 5 additional test cases, all completed successfully. Representative input images and their corresponding reconstructions for this task are shown in Figure~\ref{fig:task1-preds}.

\begin{figure}
    \centering
    \includegraphics[width=0.95\linewidth]{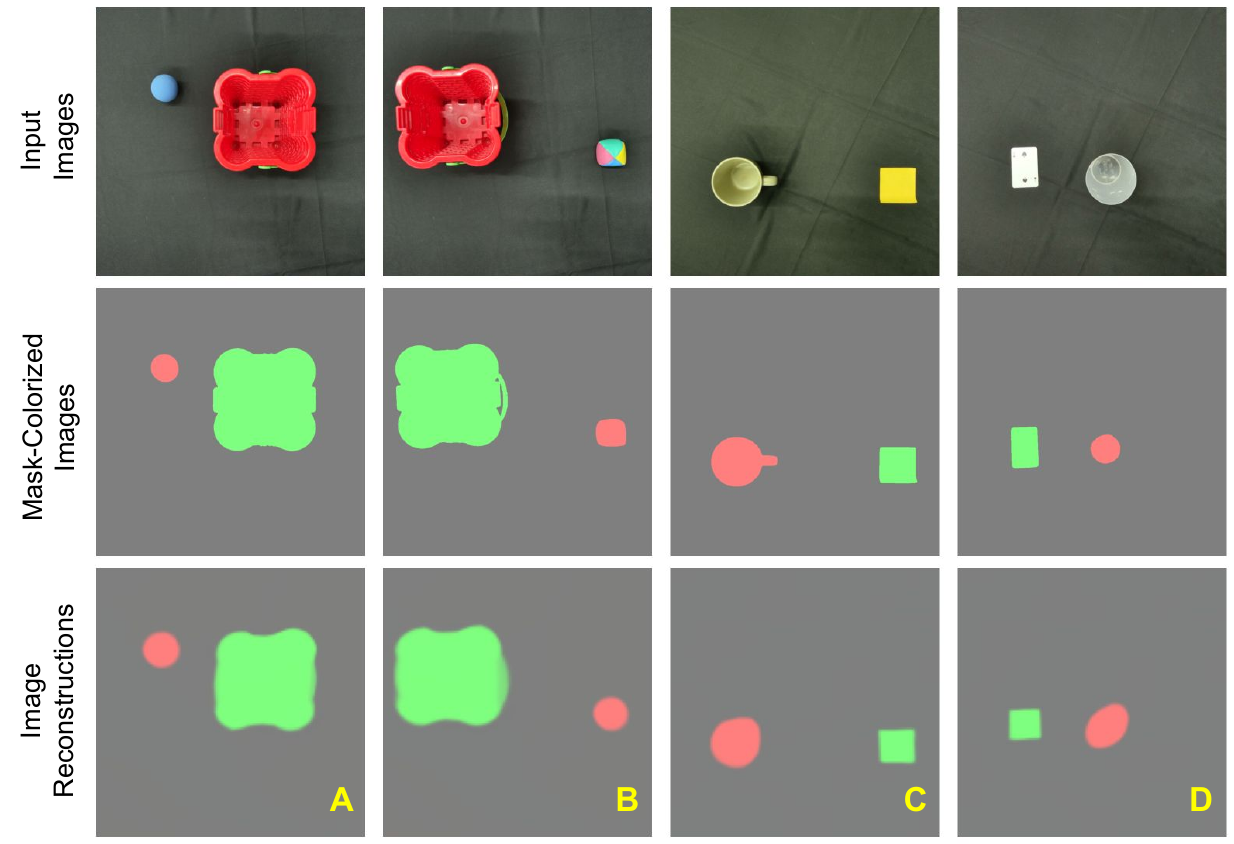}
    \caption{Illustration of the perception pipeline: input images, language-conditioned mask colorizations produced by the vision–language model, and the resulting image reconstructions.}
    \label{fig:task23-preds}
\end{figure}

The second and third tasks evaluate category-level generalization using language-guided segmentation models. In the second task, the robot picks up a ball and places it in a container; in the third, it picks up a cup and places it at a target location marked by a piece of paper. Reliable grasping requires approximately 2 cm and 1 cm horizontal accuracy, respectively. We use a language-based segmentation algorithm~\cite{medeiros2023langsam} to detect object masks. The prompts “ball” and “box” (Task 2), and “cup” and “rectangle paper” (Task 3) are used to obtain segmentation masks, which are then used to generate colorized images and object masks. During training, the same objects are used across demonstrations. The models for the second and third tasks are trained with 20 and 30 demonstrations, respectively. At test time, we additionally evaluate on unseen objects from the same semantic categories. For the third task, these include two mugs and a watering can with similar cylindrical geometry and a top opening, allowing the demonstrated grasping strategy to transfer without modification. Figure~\ref{fig:task23-preds} illustrates the perception pipeline, showing input images, language-conditioned colorized masks, and the corresponding reconstructions.

\begin{figure}

    \centering
    \includegraphics[width=\linewidth]{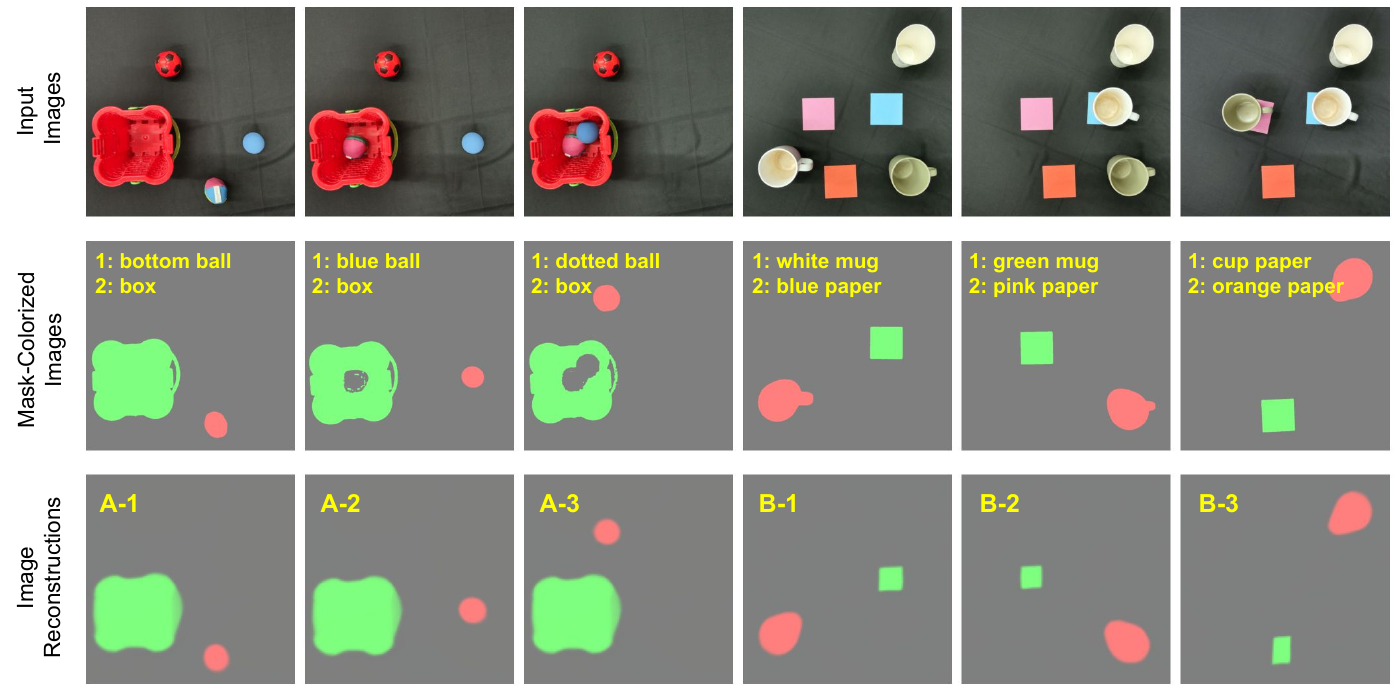}
    \caption{Example multi-step trials for the second (A-1, A-2, A-3) and third (B-1, B-2, B-3) tasks. Language prompts in the second row allow the user to specify the object to be manipulated.}
    \label{fig:task23-preds-multi}
\end{figure}

We evaluate both tasks on 15 test cases each. In the second task, the system succeeds in 14 out of 15 trials, while in the third task, it completes all test cases. The single failure in the second task occurs when testing with a previously unseen baseball, for which the model’s predicted grasp was slightly misaligned; the harder surface of the baseball reduces compliance and imposes stricter precision requirements on the grasp. Under identical test conditions, but using the training object (a tennis ball) instead of the baseball, the task succeeds, as the tennis ball’s softer, more compliant surface reduces the precision required for successful grasping. We additionally evaluate a multi-object, language-conditioned setting for both tasks. For each trial, a language prompt (e.g., “tennis ball,” “blue ball” for Task 2; “red paper,” “playing card,” “glass cup,” “mug” for Task 3) specifies the target object. These prompts are used to obtain segmentation masks and corresponding colorized images~\footnote{This evaluation assumes that the language prompt produces a correct segmentation mask for the intended target object.}. The model then generates a trajectory for the specified object. We test this over five trials per task, and in each scenario, we perform three executions—each using different prompts and colorized images. Figure~\ref{fig:task23-preds-multi} shows representative scenes, prompts, colorized images, and reconstructions used for trajectory generation. Across all trials, the system successfully completes all three pick-and-place executions.

\begin{figure}[!t]
    \centering
    \includegraphics[width=0.95\linewidth]{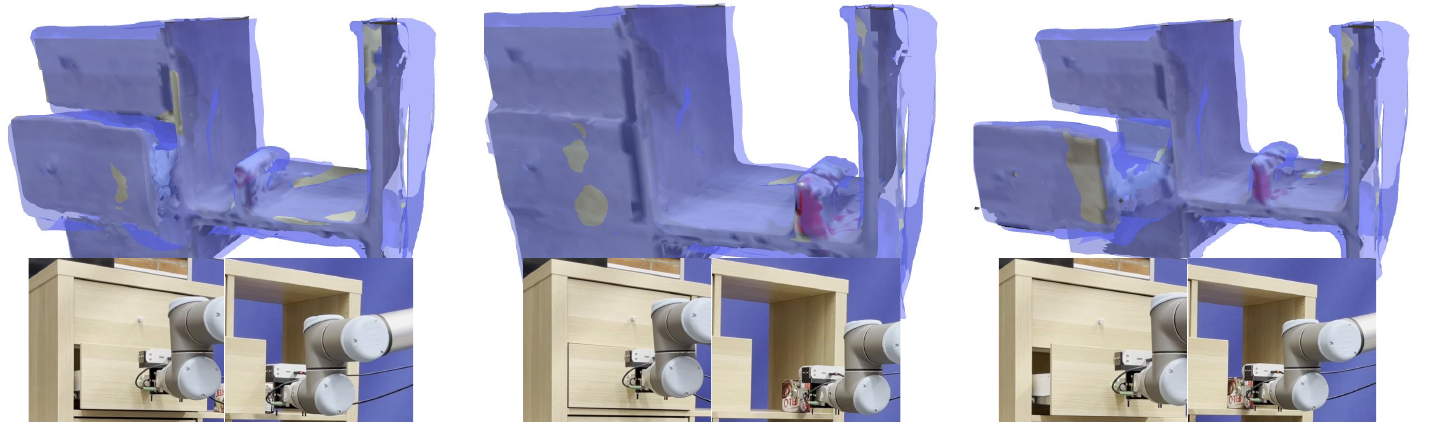}
    \caption{Input meshes and reconstructions produced by the proposed method. Snapshots from robot motion corresponding to key grasp poses are shown right below the reconstructions. The reconstructed surfaces (shown in blue) capture the drawer openings and box positions, supporting motions that enable successful grasps of the drawer handle and the box.
    }
    \label{fig:task4}
\end{figure}

In the final task, the robot places a carton box inside a drawer. This task requires opening the drawer, retrieving the box from a neighboring compartment, and placing it inside the drawer, while the initial drawer opening and the box’s initial position vary across trials. Grasping the drawer handle requires about 1 cm accuracy. This task tests the method’s ability to generalize across full three-dimensional geometry and variations in scene surfaces. To represent the scene, we scan it from several predefined camera poses and estimate a Euclidean Signed Distance Field (ESDF)~\cite{millane2024nvblox}, which is then converted into an occupancy field by marking locations with $\mathrm{ESDF} < \epsilon$ (with $\epsilon = 0.02$) as occupied. The model is trained using four such occupancy fields paired with their corresponding motion trajectories~\footnote{With fewer than four examples, the scene representation model is unable to capture the drawer’s articulation.}. Evaluation on ten novel test configurations results in successful task completion in all cases. This level of generalization from only a few demonstrations is enabled by the object-centric representation, which captures the drawer opening and box position through interpolation between two latent corner vectors for each factor (open and closed for the drawer, and leftmost and rightmost positions for the box). Figure~\ref{fig:task4} shows the meshes acquired from scene scans together with the corresponding reconstructions obtained from the predicted occupancy fields, along with snapshots from task execution. The reconstructed meshes accurately reflect the positions of both the drawer and the box. The snapshots highlight the key poses that result in successful task execution. We further evaluate the model on eight held-out demonstrations collected under the same conditions as the training data. The average position error at key drawer-handle and box-grasp poses is 0.32 cm and 0.35 cm, respectively. The measured prediction errors indicate that the learned model maintains effective generalization across previously unseen configurations. 

\section{Conclusion}
This work presents a compositional framework for motion generation directly from visual demonstrations by coupling object-centric neural representations with a temporal mixture-of-experts model. The spatial mixture-of-experts learns smooth and compact latent representations that capture object-level variability from a small number of images, while the temporal mixture-of-experts exploits these latent representations to model motion as a sequence of object-conditioned primitives. The results demonstrate that compositional modeling at the object level provides a powerful paradigm for learning from demonstration, offering interpretable latent structure, data-efficient motion generation, and systematic generalization in both simulation and real-world environments. To support broader and more flexible generalization at the scene level, the framework will be extended to incorporate neural radiance field–based scene representations. 

\bibliographystyle{IEEEtran}
\bibliography{IEEEabrv, references}

\end{document}